\pdfoutput=1
\documentclass[11pt]{article}

\usepackage[final]{coling}

\usepackage{times}
\usepackage{latexsym}

\usepackage[T1]{fontenc}
\usepackage[utf8]{inputenc}

\usepackage{microtype}

\usepackage{inconsolata}

\usepackage{graphicx}

\usepackage{paralist}
\usepackage{arydshln}

\usepackage{color}

\title{Zero-shot and Few-shot Learning with Instruction-following LLMs for Claim Matching in Automated Fact-checking}

\author{Dina Pisarevskaya \and Arkaitz Zubiaga \\
         Queen Mary University of London, UK \\ \{d.pisarevskaya, a.zubiaga\}@qmul.ac.uk}

\begin{document}
\maketitle
\begin{abstract}
The claim matching (CM) task can benefit an automated fact-checking pipeline by putting together claims that can be resolved with the same fact-check. In this work, we are the first to explore zero-shot and few-shot learning approaches to the task. We consider CM as a binary classification task and experiment with a set of instruction-following large language models (GPT-3.5-turbo, Gemini-1.5-flash, Mistral-7B-Instruct, and Llama-3-8B-Instruct), investigating prompt templates. We introduce a new CM dataset, ClaimMatch, which will be released upon acceptance. We put LLMs to the test in the CM task and find that it can be tackled by leveraging more mature yet similar tasks such as natural language inference or paraphrase detection. We also propose a pipeline for CM, which we evaluate on texts of different lengths. 
\end{abstract}

\section{Introduction}

Claim matching (CM) is the task of determining if two claims can be verified using the same piece of evidence or fact-check, which can make the automated components of the fact-checking process more efficient \citep{zeng2021automated}.~\citet{ceddeno2020} stated the problem, and~\citet{shaar-etal-2020-known} defined the task, introducing the first dataset and baselines. While some see CM as a ranking task that finds the claims most related to a given claim~\citep{shaar-etal-2020-known,shaar2021b,shaar2022,kazemi-etal-2021-claim,kazemi2022}, others tackle it as a binary classification task determining if a pair of claims should be matched to each other or not~\citep{kazemi-etal-2021-claim,kazemi2022,10.1145/3589335.3651910}.

Large language models (LLMs) have become a powerful tool in Natural Language Processing (NLP) in recent years to solve a wide variety of tasks~\citep{openai2024gpt4technicalreport}. The suitability of instruction-following LLMs to CM is yet to be studied. Finding out if LLMs can be used to determine semantic similarity (SS) between statements has recently received a great deal of attention from the NLP community~\citep{bubeck2023sparksartificialgeneralintelligence}. However, two matching claims can have low SS. A complex input claim can contain multiple sub-claims~\citep{shaar-etal-2020-known} or encompass reasoning between statements.
LLMs have shown high results without task fine-tuning but based only on hand-crafted task descriptions with instructions, and zero or limited number of task examples (prompts)~\citep{NEURIPS2020_1457c0d6}.
Zero-shot and few-shot learning approaches can be particularly useful in the context of CM, given the expensive and time-consuming nature of manually labelling data where annotators need to go through long lists of claims to find matches. Hence, in our work, we explore the effectiveness of zero- and few-shot approaches to CM with instruction-following LLMs, where state-of-the-art methods~\citep{kazemi-etal-2021-claim,kazemi2022} rely on larger training data for fine-tuning.

Given the newness of the CM task, little is known about how to approach the task: as a new task, or do approaches from close, more mature tasks suffice (e.g. natural language inference (NLI) or paraphrase detection (PD)). We investigate, for the first time, prompt-based, instruction-following LLMs for CM, in turn exploring the suitability of NLI, PD or bespoke CM prompts. We experiment on two datasets; (1) where there is a dearth of suitable datasets, we compile ClaimMatch, a new dataset extending that from CheckThat 2022~\citep{nakov-clef-2022}, for claim matching with short texts, and (2) dataset based on \citet{kazemi2022} for claim matching with long texts. 

Claim matching is a specific and complicated fact-checking task, but, with contemporary instruction-following LLMs, it can be resolved as a natural language inference or paraphrase detection task. We find that LLMs yield superior results compared to the SS baselines, and state-of-the-art classification results with fine-tuned XLM-RoBERTa. Providing few-shot examples helps LLMs better tackle the CM task. In zero-shot and few-shot setups, both PD and NLI templates are suitable for the CM task. Although different LLMs in the experiments require different prompt engineering approaches and differ if CM / PD / NLI setup is the best one for them, most of them are consistent in best prompts for CM, PD and NLI few-shot setups with a single instruction used. After running experiments with four models, we identify that Mistral with a PD template provides the best few-shot results (F1 95\%), which with only 10 few-shot train examples gets close to the fine-tuned XLM-RoBERTa (F1 96.2\%). With NLI and PD prompt templates combined, Gemini few-shot outperforms it (F1 97.2\%), leading to the best score in the comparison.

Our main contributions are:
\begin{compactitem}
 \item we investigate PD, NLI and CM prompts to address our key research question to determine if CM is a unique task;
 \item we are the first to examine the CM task with instruction-following LLMs in zero-shot and few-shot learning scenario, comparing performance against baseline ans state-of-the-art CM methods;
 \item we introduce and release ClaimMatch \footnote{\url{https://github.com/deenochka/ClaimMatch\_dataset/}}, a new dataset for claim matching;
 \item we evaluate the CM pipeline on texts of different lengths.
\end{compactitem}

\section{Related Work}

\paragraph{Claim matching.} The task can be formulated for different settings: at the short / claim level or at the longer / document level. The task can be addressed as a ranking or as a classification task. Regarding the matching of short claims with other short texts, sentence-based BERT-like embeddings cosine similarity methods for estimating textual similarity of claims at the claim level are usually used.~\citet{shaar-etal-2020-known} proposed to handle CM as a ranking task, using methods based on BM25 ranking and BERT-based cosine similarity.
Textual similarity methods were addressed further for the CM task, using pre-trained LMs (Universal Sentence Encoder, RoBERTa, sentence-BERT, etc.), for English (i.e. ~\citealp{ceddeno2020,mihailova2021,pritzkau2021}).

SS detection methods can be considered as the state-of-the-art baseline. ~\citet{shaar2021c} investigated the task from a document-level perspective: given an input document, how to identify all sentences that contain a claim that can be verified by previously fact-checked claims. State-of-the-art similarity measures and ranking measures between possible input–verified pairs were implemented: BM25, NLI Score~\citep{nie-etal-2020-adversarial}, BERTScore~\citep{zhang2020}, Sentence-BERT (SBERT) cosine similarity scores~\citep{reimers-gurevych-2019-sentence}, and SimCSE~\citep{gao-etal-2021-simcse}. For non-English data,~\citet{kazemi-etal-2021-claim} trained Indian XLM-R model, performing SS-based ranking, and classification experiments. 
To the best of our knowledge, we are the first to explore CM in zero-shot and few-shot settings, using pre-trained instruction-following LLMs. 

~\citet{kazemi2022} studied how to find matches from longer fact-checks for short claims made in tweets. The task was addressed as a ranking and as a binary classification task.
Multilingual XLM-RoBERTa pre-trained LM was fine-tuned on the dataset (with different settings: all data together or data in different languages separately).
\citet{10.1145/3589335.3651910,10.1145/3589335.3651504} utilised domain fine-tuning (based on a synthetic training set) of generative pre-trained LLMs (GPT-3.5-Turbo, Llama-2-13b-chat-hf, and Llama-2-7b-chat-hf) for CM, considering it as a RTE (recognising textual entailment) classification task with 3 classes: entailment (the truth of the first claim implies the truth of the second claim), neutral, contradiction between two claims (truth of the first claim implies that second claim is false). This task setup differs from our task setup where CM is examined as a binary classification task to check if two claims can be checked with the same piece of evidence. In our setup, matches are not necessarily related to entailment (e.g. ``the sky is blue'' and ``the sky is red'' are definite matches in our setup, while one doesn’t entail the other). 
Following the recent efforts, we also tackle CM as a binary classification task in hitherto unexplored zero-shot and few-shot settings, studying it for shorter and for longer texts.

\paragraph{Zero-shot and few-shot learning}
Let models perform on new tasks without fine-tuning. Zero-shot and few-shot setups assume addition of extra prompt text to each input example, so a model can learn the task based on such train examples and instructions~\citep{schick-schutze-2021-just,le-scao-rush-2021-many} and provide the answer that corresponds to a final label, filling the unfilled slots in a prompt. This method depends on manually or automatically creating better label words and templates for prompts (prompt engineering). As some prompts work better than others, methods to automatically generate high-quality and diverse prompts have been suggested~\citep{jiang-etal-2020-know,eb6d59f260f0477db529155cc08d3e61,lu-etal-2022-fantastically,lester-etal-2021-power,qin-eisner-2021-learning,li-liang-2021-prefix}.
~\citet{schick-schutze-2021-exploiting} studied how to detect appropriate label words automatically, but the hand-picked ones performed better, while~\citet{gao-etal-2021-making} applied prompt generation with automated selection of both templates and label words, that achieved comparable results to hand-crafted prompts.
~\citet{webson-pavlick-2022-prompt} evaluated different manually written prompt templates for NLI task. %However,
~\citet{logan-iv-etal-2022-cutting} showed that tuning only the prompt yields worse performance than with manually designed prompts, 
but suggested that it might depend on the models and settings. Promptsource~\citep{bach-etal-2022-promptsource} provides a variety of prompts for zero-shot and few-shot learning, including prompts for PD and NLI. 

\paragraph{LLMs applications.} Recent advances with LLMs~\citep{dang2022promptopportunitieschallengeszero} have led to their exploration across various tasks. Prompt-based approaches can be used for classification tasks in zero-shot and few-shot setups: NLI and SS as a classification task~\citep{gu-etal-2022-ppt}, topic classification~\citep{hu-etal-2022-knowledgeable}, and paragraph classification~\citep{deng-etal-2022-beike,wang-etal-2022-pingan}.~\citet{wangentalment2021} reformulated NLP tasks as the entailment ones, and then fine-tuned the model with as little as 8 examples. Reasoning abilities of recent LLMs were yielded in~\citep{NEURIPS2022_9d560961,zhou2023leasttomostpromptingenablescomplex,press-etal-2023-measuring,wei-jie-etal-2024-interpretable}, encouraging their use in tasks that require commonsense reasoning and complex language understanding, such as multiple fact-checking tasks: zero-shot scientific claims verification with GPT-3.5 and GPT-4~\citep{alvarez-etal-2024-zero}, generating fact-checking explanations with GPT-4, Claude 2, and PaLM 2~\citep{hsu-etal-2024-enhancing}, creating claim-focused summaries in zero-shot and few-shot setups with GPT-3.5 for claim verification~\citep{chen-etal-2024-complex}, 
or exploring few-shot scenarios with GPT-3.5 to verify claims with a knowledge source~\citep{li2024selfcheckerplugandplaymodulesfactchecking}. 
In this work, we study the CM task, using instruction-following, prompt-based LLMs for it.

\section{Datasets}

We use two datasets in our research: ClaimMatch, which we create by extending and adapting data from \citet{nakov-clef-2022} to experiment with short texts; and longer texts (LT) dataset based on \citet{kazemi2022} to experiment with long texts.

\textbf{Short texts.} To create the ClaimMatch dataset, we rely on texts from the fully available public dataset~\citep{nakov-clef-2022}, which was in turn created by building on and extending \citet{shaar-etal-2020-known}.
We use the multi-domain English subset of the data and get rid of the remainder data pertaining to the political domain only.
It consists of tweets as input claims and previously verified claims from the corresponding articles from the Snopes website. It contains 1,398 claim pairs (training, development and dev-test sets combined), and 14,245 previously checked claims in general.  

We extend this data as it originally contained only positive cases of claim matches, with no negative cases. For positive class examples, we chose 500 claim pairs from the dataset. For the negative class examples, we created them: for each of the selected input claims, we took another verified claim not from their pair (each verified claim from the dataset could be used only once). We used such new pairs as negative examples.\footnote{We manually double-checked all generated pairs to prevent false positives or false negatives.} Table \ref{tab:claimmatch} in the appendix shows samples of positive and negative claim matches.

All texts were preprocessed: urls, retweet markers (``RT'') and emojis\footnote{Emojis were processed with the package https://pypi.org/project/emoji/} were removed. Username (@) and hashtag (\#) markers were removed while leaving the mentions, to keep the content of short texts. Text of a verified claim includes its title, subtitle and main text, taking into account the importance of the titles of articles about claims in previous research~\citep{10.1007/978-3-030-99736-6_25}.

Our resulting test set, used in the LLMs experiments, contains 1,000 examples: 500 for the positive class and 500 for the negative class. Both input claims and verified claims are short. For the input claims, the average length after preprocessing is 194 characters (39 tokens). For verified claims, average length after preprocessing is 303 characters (56 tokens). This new dataset is specified for CM as a binary classification task, and will be released. 

\textbf{Longer texts.} We use the Kazemi dataset~\citep{kazemi2022}) for domain transfer evaluation on longer texts. It is based on fact-checks obtained from several sources, and contains 5,028 pairs of long texts \& short claims, of which we firstly sample 1,000 due to computational restrictions. The preprocessing of texts from this dataset is the same as for the first one. In addition, we remove pairs with Levenshtein distance similarity ratio between a text and a claim more than 80\%, to remove (near-)duplicates. The average lengths after preprocessing are: 209 characters (41 token) for claims, 3,776 characters (729 tokens) for articles. To create the LT dataset, we take a sample of 129 positive examples and create 129 negative examples with the same procedure as described for short texts above (258 test examples in general). The dataset~\citep{kazemi2022} is public, but its usage required collecting twitter data to build connections between texts. As the dataset was not publicly available online, we assume that it was not used for pre-training LLMs, which is important while using public datasets for evaluation with LLMs. Hence, the LT dataset is suitable for the evaluation of the proposed CM approach with LLMs. 

\section{Experimental Setup}

We use four popular instruction-following LLMs for prompt-based experiments, including 2 models with paid API, and 2 freely available models (accessed by request):

\begin{compactenum}
 \item Gpt-3.5-turbo-0125 (\textbf{GPT-3.5}): the latest and cost effective model in the GPT-3.5 family
 allows instruction following and works for traditional completions non-chat tasks as well\footnote{\url{https://platform.openai.com/docs/models/gpt-3-5}}. Tokens context length was limited by the model context length.

 \item Gemini-1.5-flash (\textbf{Gemini}): the fastest and cheapest model from the Gemini family\footnote{\url{https://blog.google/technology/ai/google-gemini-update-flash-ai-assistant-io-2024/}} allows instruction following. Default context length was used.

 \item Mistral-7B-Instruct-v0.3 (\textbf{Mistral}): the latest instruct fine-tuned version of Mistral-7B-v0.3\footnote{\url{https://huggingface.co/mistralai/Mistral-7B-Instruct-v0.3}}, the popular example of open-source LLMs of such size. Maximal new tokens is set to 400, as it places the full model's answers. 

 \item Llama-3-8B-Instruct (\textbf{Llama}): the last open-access Llama model of such size\footnote{\url{https://huggingface.co/meta-llama/Meta-Llama-3-8B-Instruct}}. The hyperparameters were: temperature 0.6, top\_p 0.9, maximum of new tokens 400. 
\end{compactenum}

The latter two models were set up to fit the Colab Pro+ resources (A100 GPU with 40 GB RAM). For all four models, their zero-shot and few-shot setups were examined with default system instruction \& specified user instruction (``single instruction''), or with specified system instruction \& specified user instruction (``ensemble instruction''). 

We list the selected CM,
PD and NLI prompt template sets for zero-shot and few-shot learning in Table \ref{tab:templates}. CM templates were selected after empirical tests from a broader initial list of CM templates. PD and NLI template sets are based on PromptSource base~\citep{bach-etal-2022-promptsource}.
For Mistral and Llama, questions were moved to the beginning of the template to adhere to model requirements.

\begin{table*}[htb]
    \centering
    \small
    \begin{tabular}{ll}
     \hline
     \multicolumn{2}{c}{\textbf{CM templates}} \\
     \hline
     \textbf{CM-1} & A Matches to B. Correct? Answer: [yes/no] \\
     \textbf{CM-2} & A Means that B. Correct? Answer: [yes/no] \\
     \hline
     \hline
     \multicolumn{2}{c}{\textbf{PD templates}} \\
     \hline
     \textbf{PD-1} & A. B. Question: Do A and B express the same meaning? Yes or no? Answer: [yes/no] \\
     \textbf{PD-2} & A. B. Question: Do A and B express the same meaning? Answer: [yes/no] \\
     \textbf{PD-3} & A. B. Question: Do A and B have similar meanings? Yes or no? Answer: [yes/no] \\
     \textbf{PD-4} & A. B. Question: Are A and B saying the same thing? Yes or no? Answer: [yes/no] \\
     \textbf{PD-5} & A. B. Question: Are A and B essentially the same? Yes or no? Answer: [yes/no] \\
     \textbf{PD-6} & A. B. Question: Do A and B both refer to the same event? Yes or no? Answer: [yes/no] \\
     \hline
     \hline
     \multicolumn{2}{c}{\textbf{NLI templates}} \\
     \hline
     \textbf{NLI-1} & Suppose it's true that A. Then, is B. Question: Is true or false? Answer: [true/false] \\
     \textbf{NLI-2} & Take the following as truth: A. Then B is true or false? Answer: [true/false] \\
     \textbf{NLI-3} & A. Based on the previous statement, is it true that B? Yes or no? Answer: [yes/no] \\
     \textbf{NLI-4} & Given A Is it guaranteed true that B? Yes or no? Answer: [yes/no] \\
     \textbf{NLI-5} & Suppose A. Can we infer that B? Yes or no? Answer: [yes/no] \\
     \hline
    \end{tabular}
    \caption{Templates used for claim matching (CM), paraphrase detection (PD) and natural language inference (NLI).}
    \label{tab:templates}
\end{table*}

In zero-shot settings, only templates were given to models. For few-shot, we used the priming method (in-context learning): together with a template, labeled examples are included in an input sequence, so a model makes a prediction~\citep{NEURIPS2020_1457c0d6}. We chose a sample of 5 positive examples from the data. As negative examples, we took 5 random input claims and selected for them 5 verified claims not from their pairs. All these claims were not used in the test set. So the set of train examples was balanced. Based on a prompt with these selected examples, the models were able to produce binary classification labels for the new examples. We used  ``yes''/``no'' (templates excepting NLI 1-2) or ``true''/``false'' (NLI 1-2 templates) as label words for the task, for two corresponding classes: ``match'' (positive)/``not match'' (negative). The final order of positive and negative examples was mixed.

In addition to the instruction-following LLMs, we include in the evaluation: 
\begin{compactenum}
\item State-of-the-art model (SOTA): multilingual XLM-RoBERTa-base\footnote{\url{https://huggingface.co/docs/transformers/en/model\_doc/xlm-roberta}} (\textbf{XLM-R}) model that, after fine-tuning on data, showed the best results for CM as classification task in~\citep{kazemi2022}. In our study, it was fine-tuned for the CM task on the dataset of remaining 1,790 claim pairs, not included in test set (10 epochs, learning rate 1e-5). 
\item Baselines: (1) SS-based model using All-MiniLM-L6-v2\footnote{sentence similarity model with the highest number of downloads on huggingface: https://huggingface.co/sentence-transformers/all-MiniLM-L6-v2} (\textbf{All-MiniLM}) and (2) SS-based model using text-embedding-3-small\footnote{\url{https://platform.openai.com/docs/guides/embeddings/embedding-models}} (\textbf{embedding3}).
\end{compactenum}

As SS-based methods need a threshold to determine the class, a separate validation set of 1,000 claim pairs was used to calculate this threshold.
Median SS score for the positive class was chosen as a threshold (0.64 for All-MiniLM and 0.63 for embedding3). In the test set, all claim pairs with SS >= this threshold got positive class labels. 

We used the standard evaluation metrics for classification: F1 score (weighted) and accuracy (in Error analysis, precision (weighted) and recall (weighted) are added). For all models, less than 5\% examples did not obtain a clear classification label (``partial match'' in a longer model output), so they were included in the negative class.

\section{Zero-shot and few-shot experiments}

To identify the most effective user instructions for a prompt, we investigate different templates on ClaimMatch and compare results in Table \ref{tab:results}.

\begin{table*}[htb]
    \centering
    \small
    \begin{tabular}{lccclccclccc}
     \cline{1-7}
     \cline{9-11}
     \multicolumn{7}{c}{\textbf{FEW-SHOT}} & & \multicolumn{3}{c}{\textbf{ZERO-SHOT}} \\
     \cline{1-7}
     \cline{9-11}

     \cline{1-3}
     \cline{5-7}
     \cline{9-11}
     Model & F1, \% & Acc., \% & & Model & F1, \% & Acc., \% & & Model & F1, \% & Acc., \% \\
     \cline{1-3}
     \cline{5-7}
     \cline{9-11}
     \multicolumn{3}{c}{\textbf{PD templates}} & & \multicolumn{3}{c}{\textbf{NLI templates}} & & \\
     \cline{1-3}
     \cline{5-7}
     \multicolumn{3}{c}{\textbf{PD-1}} & & \multicolumn{3}{c}{\textbf{NLI-1}} & & \multicolumn{3}{c}{\textbf{PD-6}} \\
     %\hline
     GPT-3.5 & 80.3 & 81.0 & & GPT-3.5 & 70.6 & 70.9 & & GPT-3.5 & \textbf{89.2} & \textbf{89.3} \\     
     Gemini & 80.5 & 81.2 & & Gemini & 66.2 & 69.3 & & Gemini & \textbf{90.7} & \textbf{90.7} \\
     Mistral & 93.3 & 93.3 & & Mistral & 85.5 & 85.7 & & Mistral & 85.6 & 85.9 \\
     Llama & 52.4 & 59.7 & & Llama & 54.9 & 58.1 & & Llama & \textbf{88.0} & \textbf{88.1} \\
     %\hline
     \cdashline{1-3}
     \cdashline{5-7}
     \cdashline{9-11}
     \multicolumn{3}{c}{\textbf{PD-2}} & & \multicolumn{3}{c}{\textbf{NLI-2}} & & \multicolumn{3}{c}{\textbf{NLI-5}} \\
     %\hline
     GPT-3.5 & 74.2 & 75.7 & & GPT-3.5 & 67.5 & 67.8 & & GPT-3.5 & 70.6 & 72.7 \\
     Gemini & 77.2 & 78.3 & & Gemini & 61.4 & 65.3 & & Gemini & 71.0 & 73.0 \\
     Mistral & 92.7 & 92.7 & & Mistral & 84.1 & 84.2 & & Mistral & 79.5 & 80.3 \\
     Llama & 64.9 & 67.2 & & Llama & 36.2 & 50.2 & & Llama & 74.3 & 75.7 \\
     %\hline
     \cdashline{1-3}
     \cdashline{5-7}
     \cdashline{9-11}
     \multicolumn{3}{c}{\textbf{PD-3}} & & \multicolumn{3}{c}{\textbf{NLI-3}} & & \multicolumn{3}{c}{\textbf{CM-1}} \\
     %\hline
     GPT-3.5 & 74.2 & 75.7 & & GPT-3.5 & 69.8 & 70.8 & & GPT-3.5 & 69.9 & 71.7 \\
     Gemini & 85.3 & 85.6 & & Gemini & 74.9 & 76.1 & & Gemini & 62.8 & 63.3 \\
     Mistral & 93.7 & 93.7 & & Mistral & 78.2 & 78.4 & & Mistral & \textbf{88.9} & \textbf{89.0} \\
     Llama & 72.9 & 72.9 & & Llama & 86.4 & 86.5 & & Llama & 70.9 & 72.9 \\
     %\hline
     \cdashline{1-3}
     \cdashline{5-7}
     \cline{9-11}
     \multicolumn{3}{c}{\textbf{PD-4}} & & \multicolumn{3}{c}{\textbf{NLI-4}} & & \\
     %\hline
     GPT-3.5 & 78.4 & 79.3 & & GPT-3.5 & 63.8 & 64.2 & &  \\
     Gemini & 80.2 & 80.9 & & Gemini & 77.9 & 78.7 & & \\
     Mistral & 93.5 & 93.5 & & Mistral & 79.1 & 79.2 & &  \\
     Llama & \textbf{78.4} & \textbf{78.4} & & Llama & \textbf{91.8} & \textbf{91.8} & &  \\
     %\hline
     \cdashline{1-3}
     \cdashline{5-7}
     \cline{9-11}
     \multicolumn{3}{c}{\textbf{PD-5}} & & \multicolumn{3}{c}{\textbf{NLI-5}} & & \multicolumn{3}{c}{\textbf{MANY-SHOT}} \\
     \cline{9-11}
     %\hline
     GPT-3.5 & 78.5 & 79.4 & & GPT-3.5 & \textbf{75.7} & \textbf{76.3} & & \multicolumn{3}{c}{SOTA model} \\
     Gemini & 85.5 & 85.8 & & Gemini & \textbf{93.4} & \textbf{93.4} & & XLM-R & \textbf{96.2} & \textbf{96.2} \\
     Mistral & 94.2 & 94.2 & & Mistral & \textbf{88.3} & \textbf{88.3} & & \multicolumn{3}{c}{Baselines}  \\
     Llama & 66.2 & 67.7 & & Llama & 52.8 & 59.7 & & All-MiniLM & 76.6 & 77.8 \\        
     %\hline
     \cdashline{1-3}
     \cline{5-7}
     \multicolumn{3}{c}{\textbf{PD-6}} & &  & & & & embedding3 & \textbf{77.2} & \textbf{78.3} \\
     \cline{9-11}
     %\hline
     GPT-3.5 & \textbf{84.8} & \textbf{85.1} & & \multicolumn{3}{c}{\textbf{CM templates}} & & & & \\
     \cline{5-7}
     Gemini & \textbf{90.3} & \textbf{90.4} & & \multicolumn{3}{c}{\textbf{CM-1}} & & &\\
     Mistral & \textbf{95.0} & \textbf{95.0} & & GPT-3.5 & \textbf{75.8} & \textbf{76.9} & &\\
     Llama & 60.0 & 64.5 & & Gemini & \textbf{72.2} & \textbf{73.8} & & & &\\
     \cline{1-3}
      & & & & Mistral & \textbf{90.6} & \textbf{90.6} & & & & \\
      & & & & Llama & 77.6 & 78.3 & & & & \\
      \cdashline{5-7}
      & & & & \multicolumn{3}{c}{\textbf{CM-2}} & & & & \\
      & & & & GPT-3.5 & 71.1 & 72.7 & & & & \\
      & & & & Gemini & 67.2 & 69.8 & & & & \\
      & & & & Mistral & 89.9 & 89.9 & & & & \\
      & & & & Llama & \textbf{81.7} & \textbf{81.8} & & & & \\
    \cline{5-7}
    \end{tabular}
    \caption{Few-shot (left) and zero-shot (top right) performance with PD, NLI and CM templates. Many-shot SOTA and baselines (bottom right).}
    \label{tab:results}
\end{table*}

\begin{table*}[htb]
    \centering
    \small
    \begin{tabular}{lccclcc}
     \cline{1-3}
     \cline{5-7}
     \multicolumn{3}{c}{\textbf{FEW-SHOT}} & & \multicolumn{3}{c}{\textbf{ZERO-SHOT}} \\
     \cline{1-3}
     \cline{5-7}
     Model & F1, \% & Acc., \% & & Model & F1, \% & Acc., \% \\
     \cline{1-3}
     \cline{5-7}     
     \multicolumn{3}{c}{\textbf{CM-1 \& PD-6 template}} & & \multicolumn{3}{c}{\textbf{CM-1 \& PD-6 template}} \\
     GPT-3.5 fs & 40.8 & 69.0 & & GPT-3.5 zs & 95.7 & 95.7 \\
     Gemini fs & 91.1 & 91.2 & & Gemini zs & \textbf{97.1} & \textbf{97.1} \\
     Mistral fs & 92.3 & 92.3 & & Mistral zs & \textbf{96.2} & \textbf{96.2} \\
     Llama fs & 66.0 & 69.2 & & Llama zs & 91.9 & 91.9 \\
     \cdashline{1-3}
     \cdashline{5-7}
     \multicolumn{3}{c}{\textbf{PD-6 \& PD-6 template}} & & \multicolumn{3}{c}{\textbf{PD-6 \& PD-6 template}} \\
     GPT-3.5 fs & 87.8 & 87.9 & & GPT-3.5 zs & 95.6 & 95.6 \\
     Gemini fs & 95.1 & 95.1 & & Gemini zs & 96.0 & 96.0 \\
     Mistral fs & 94.9 & 94.9 & & Mistral zs & 95.5 & 95.5 \\
     Llama fs & 60.7 & 65.2 & & Llama zs & 95.6 & 95.6 \\
     \cdashline{1-3}
     \cdashline{5-7}     
     \multicolumn{3}{c}{\textbf{NLI-5 \& PD-6 template}} & & \multicolumn{3}{c}{\textbf{NLI-5 \& PD-6 template}} \\
     GPT-3.5 fs & 59.5 & 59.9 & & GPT-3.5 zs & \textbf{96.1} & \textbf{96.1} \\ 
     Gemini fs & \textbf{97.2} & \textbf{97.2} & & Gemini zs & 97.1 & 97.1 \\
     Mistral fs & 89.0 & 89.0 & & Mistral zs & 92.5 & 92.5 \\
     Llama fs & 48.0 & 57.4 & & Llama zs & 94.7 & 94.7 \\ 
     \cline{1-3}
     \cline{5-7}     
    \end{tabular}
    \caption{Few-shot (left) and zero-shot (right) performance with ensemble instructions.}
    \label{tab:instresults}
\end{table*}

\textbf{Few-shot results with a ``single instruction''}. Overall, our results show that PD and NLI templates outperform CM templates, suggesting that reformulating the CM task as a PD and NLI helps. The improvement is primarily noticeable with PD, suggesting that, with the datasets at hand, matching claims can be handled as paraphrases. This is however not consistent; certain templates such as PD-6 and NLI-5 can outperform CM templates, but not all PD and NLI templates do.

Different LLMs require different prompt engineering approaches and differ if CM / PD / NLI setup is the best one for them, but most of them are consistent in best prompts within each of the CM, PD and NLI setups: for GPT-3.5, Gemini and Mistral, templates CM-1, PD-6 and NLI-5 are the most effective for their sets. Only for Llama, templates CM-2, PD-4 and NLI-4 are better. For GPT-3.5
and Mistral, PD-6 template performs better, for Gemini - NLI-5, for Llama - NLI-4.

Prompt templates in the same set can have similar meanings yet provide distinct results: for example, PD-1 and PD-2 differ slightly, but all the model results yield significant differences. The results of the models also vary in different ways in the template sets and among the models and the templates. Wording-dependent prompt engineering separately for each model is required here. Mistral yields the highest overall metrics in all template sets. For this model, PD templates show highest scores, and the CM templates work better than the NLI templates. Llama has the lowest scores, and its results with the NLI-2 template demonstrate that the model understood the task wrongly (as a claim verification rather than as a CM task). GPT-3.5 and Gemini are not so consistent along results for template sets. Although Llama is very good with NLI-4, it performs much worse with NLI-5 where explicit inference understanding is needed. Mistral shows high consistency and impressive scores along PD templates results, but for NLI they vary much. These examples demonstrate that there are some variation across LLMs, and  one should be careful choosing the template type depending on the model. 

As for SS baselines, Mistral and Llama outperform All-MiniLM and embedding3 in both CM templates, although GPT-3.5 and Gemini results are below the baselines. With most of the PD templates, GPT-3.5, Gemini and Mistral yield higher results than both baselines, for Llama only one template works better than them. GPT-3.5 does not outperform the baselines with all NLI templates, but two NLI templates work for Gemini and Llama (with NLI-4 for both of them) better than the baselines. All Mistral NLI templates outperform the baselines. The best few-shot results of PD-6 template with Mistral (F1 95.0\%) and NLI-5 template with Gemini (F1 93.4\%) do not outperform state-of-the-art classification results with fine-tuned XLM-R (F1 96.2\%), but can be compatible with them in real-word scenarios, where only little training data are available, not sufficient for models fine-tuning.

\textbf{Zero-shot results with a ``single instruction''}. For comparison, we present zero-shot results for three templates, that showed the best results for most of the models: CM-1, PD-6, and NLI-5. As expected, for most of the settings, few-shot provides higher scores than zero-shot, and we need some examples to let LLMs better tackle the task. However, With PD-6 template, for GPT-3.5 zero-shot is better, and for Gemini - slightly better than few-shot. GPT-3.5 provides F1 89.2\%, that is still below the XLM-R F1 96.2\%. For Llama, zero-shot provides significantly higher scores with PD-6 and NLI-5 templates. It can be explained that this model gets distracted by the content of few-shot examples specifically in the given templates. Zero-shot results show the potential of tackling the CM task with PD templates, and it also shows that using just a few shots can lead to a significant gain generally, so having a few samples to train the model can improve performance.

\textbf{``Ensemble'' instructions}. As almost all models have high performance with PD-6 template, we added it specifically to their system instructions, so each model setting has 2 instructions combined: PD-6 template in system instruction and another template in user instruction. Results are presented 
in Table \ref{tab:instresults} for CM-1, PD-6 and NLI-5 templates. In all settings, excluding Gemini NLI-5, zero-shot worked much better than few-shot. Only for Llama, ``double'' PD-6 template with zero-shot is better than CM-1 and NLI-5 templates. ``Ensemble'' instructions let Gemini improve results - both for zero-shot and few-shot setups. Compared to best few-shot results with default system instructions, zero-shot results for all models are significantly better, and more similar along all models and templates. It is unexpected, because with a ``single'' instruction few-shot was in general better than zero-shot. It can be explained that models can be distracted by few-shot examples provided in user instruction only for one of two templates. In zero-shot, Mistral with CM-1 \& PD-6 template reaches state-of-the-art scores (F1 96.2\%), and Gemini outperforms them (F1 97.1\%). While CM-1 \& PD-6 template is the best zero-shot one for Mistral and NLI-5 \& PD-6 template is the best zero-shot one for GPT-3.5, Gemini, however, provides the best overall scores (F1 97.2\%) in the comparison with its few-shot NLI-5 \& PD-6 template (outperforming the state-of-the-art scores with both zero-shot and few-shot NLI-5). So few-shot training data, and templates from close tasks, such as NLI and PD, are still helpful to achieve better results. Combination of templates leads to improved and more consistent results, which make it more reliable. While PD templates were very good alone, it shows that they can be further improved by using other templates, including a CM template. For most of the cases, combinations of templates yields better scores than these two templates separately, both for few-shot and zero-shot. 

Both PD and NLI templates, as templates from close NLP tasks, seem to be appropriate for the CM task in zero-shot and few-shot setting with instruction-following LLMs. But they should be carefully examined specifically for each model, as they are more or less suitable for different models, and even similar templates let get different classification scores. Combinations of templates and choice of a zero-shot/few-shot setup should be carefully investigated, to yield better results.

\textbf{Longer texts}. We next created a pipeline for few-shot experiments for different domains and tested it on longer texts devoted to various news topics from LT dataset. Two setups were used in these experiments: Mistral few-shot PD-6 (as the best one with ``single instructions'') and Gemini few-shot NLI-5 \& PD-6 (as the best one with ``ensemble'' instructions). For both models, default context length was used. Despite~\citet{kazemi2022}, longer texts were processed on the entire text level and not on paragraph level (due to LLMs context capabilities). We experimented with two options: 1. using train examples, the same as in experiments on short texts; 2. using a sample of new 10 domain train examples from this dataset~\citep{kazemi2022}. Mistral setup resulted in 83.7\% (both F1 and accuracy scores) with short train examples, but domain train examples led to only F1 67.7\% and accuracy 70.5\%. Gemini setup yielded 98.4\% (both F1 and accuracy scores) with short examples, and 95.7\% (both F1 and accuracy scores) with domain examples. As the performance for both Mistral and Gemini models was better with domain-independent train examples (short texts) than with domain train examples, we suggest using data-independent train examples in the pipeline. Hence, in the proposed pipeline, firstly data-independent train examples should be chosen (manually or in automated way). The prompt template should be selected. Then, for each new test example, few-shot learning technique should be applied: 10 train examples should be given to a LLM, followed by the test example. Such domain transfer methods would be helpful for benchmarking and evaluation of the few-shot learning with instruction-based LLMs for CM task, on different domains and on texts of various lengths.

\section{Error Analysis}

We studied all model outputs, to deeper understand performance and why different models perform better with different prompts. For all templates except for NLI 3-5, in most of the cases the answers were accompanied with model explanations. Gemini few-shots provided more structured answers (starting with the class label, e.g. ``yes''/``no''), with less explanations than other models. 

\textbf{Fact-checking instead of CM}. ``Single instruction'' CM templates work for Mistral much better than for other models, and it understood the task very well (due to the model’s argumentation). But NLI 1-2 templates were problematic: the model did not understand the task and incorrectly addressed it as a fact-checking task, due to ``true''/``false'' in instruction question. For other models, such cases were also present for these tasks, so ``true''/``false'' questions should be excluded for CM.
NLI 3-4 templates did not cause as many fact-checking answers: they contain ``true'', but focus on two claims inference is expressed more clearly. 

\textbf{Focus on train examples} can also lead to errors. For Llama, ``single instruction'' templates performed worse than for other models (excluding NLI-4). In few-shot, it could be confused with train examples and start comparing two test claims not to each other, but to train claims (e.g. in NLI-5 few-shot recall for negative class was 21.0\%). But its zero-shot with ``ensemble'' instructions got results compatible to other models. This model requires a specific prompt engineering (e.g. more explicit numeration in train and test examples).    

\textbf{PD and NLI templates require some clarification for CM task}. Even PD-6, one of the best templates, can still be too strict for CM. In false negative answers of all models, two claims were considered by models as referring to different events, if they vary in some details. But such claims can still match each other according to the gold label. 

This error type is more specific for zero-shot settings. All four models can detect if two statements in a claim pair vary in some non-substantial or significant details. But the level of this difference, and the final class label verdict of a model are handled individually for each concrete example, model and template setup. For example, Mistral's output with CM-1 template included 36 gold label positive class examples where the model highlighted that claims in a pair are about similar, but not the same events (e.g. ``While Statement 1 does suggest that Joe Biden owns the largest mansion in his state, it does not definitively establish this fact. Therefore, it does not match Statement 2, which claims that the information is definitively known''). The model provided the positive class label only for 11 examples. But if we consider all 36 examples as positives, the final classification performance is improved: both F1 score and accuracy are 91.5\% (instead of 88.9\% and 89.0\% respectively). Hence, templates can be modified for the task. Separate model explanations should be a step of the pipeline. If a model answers that two claims are about the same topic and event, but vary in not substantial details, the pair should be processed as positive during post-processing.

\textbf{Claim specifics} can also cause errors.  The models in different setups classified one example as false negative: two claims mentioned different numbers. E.g. Llama's output with PD-6 template, after providing the negative class label: ``Statement 1 mentions that cameras on the M1 and M25 go live, and anyone going over 70 mph gets an instant ticket. This suggests that the cameras were activated with a speed limit of 70 mph. Statement 2 mentions that all speed cameras on the M1 and M25 were activated in January 2019 with a uniform speed limit of 72 miles per hour. The speed limit is different in the two statements, which indicates that they do not refer to the same event''. The discrepancies, highlighted by the model, really take place in the texts, so it can be explained why two claims can be considered as not fully matching and not verifiable by the same piece of evidence. 

\textbf{Complex claims and nested sub-claims}. Addressing this challenging CM issue, all models demonstrated understanding of sub-claims. However, with PD templates 1-5 models can answer that statements A and B are about similar, but not the same event, as B includes another sub-claim - information that is not present in A; or the same persons and locations are mentioned, but the context is different. This issue can be solved by exploring prompt templates and decomposing the CM task. 

We conducted a fine-grained study of 10 claim pairs, where at least one of the texts is a complex claim with several sub-claims in it, e.g.: ``1) In the 1640’s the Dutch inhabitants of New Amsterdam built a 12' wall to keep the bad hombres out. In 1664 the British ignored the wall and took New Amsterdam by sea. It’s now called New York. They took down the wall and built a street. It’s called Wall Street. — Joe Delmonaco (JoeDelmonaco) January 7, 2019	2) Was Wall Street Originally the Site of a ``Border Wall'' Meant to Protect New Amsterdam? Social media memes compared a defensive wall built along the northern border of New Amsterdam during the 17th century to President Donald Trump's border wall. In the 17th century, New Amsterdam built a protective wall along its northern perimeter (analogous to Trump's border wall) to keep ``bad hombres'' out, but it failed to achieve its stated purpose in that the British successfully invaded the city by sea.''	

We did not find a consistently best PD template, but found some patterns. With PD-1 and PD-2 templates, Llama and GPT-3.5 have difficulties with the provided example. Llama with PD-2 template correctly noticed that there are several sub-claims in Statement 2. It justified the negative class label for this claim pair: ``No, Statement 1 and Statement 2 do not express the same meaning. Statement 1 discusses the construction of a wall in New Amsterdam in the 1640s and its failure to prevent the British from taking the city by sea. Statement 2 compares the wall built by New Amsterdam to President Trump's border wall and states that it failed to achieve its purpose, citing the British invasion of the city.'' In general, CM for complex claims is well solved by the models. Compared to few-shot results with a ``single instruction'', all four models with ``ensemble instructions'' perform better. It shows that different instructions help a model understand the task from different perspectives. 

\section{Discussion \& Conclusion}

We propose a novel direction to tackle claim matching as a classification task: prompt-based zero-shot and few-shot learning for pre-trained instruction-following LLMs, suitable when there is no or limited train data. We create the ClaimMatch dataset with short texts and perform extensive experiments on it. We use manually created prompt templates for zero-shot and few-shot (10 train examples) settings, aimed specifically for CM or for two similar tasks: PD and NLI. We investigate if the latter two tasks can help tackle CM task. Different LLMs from our benchmark comparison require different prompt engineering approaches, but in general PD and NLI work better than CM templates.

Our results help better understand that zero-shot and few-shot methods with templates crafted for PD and NLI tasks are appropriate for CM. With instruction-following LLMs, it can be resolved on the existing data with using NLI and PD tasks approaches. Few-shot training examples are still needed to experiment with the better settings. Hence, advanced prompt engineering methods should be further carefully developed for CM. We suggest the pipeline for zero-shot and few-shot CM,
and test it with best prompt templates on longer texts. Our work makes the first contribution of prompt-based zero-shot and few-shot learning with instruction-based LLMs for CM, and provides possible further research questions and directions.

The creation of more datasets for claim matching will greatly benefit further development of the task. One possible reason why PD templates are well suited to the CM task is that existing datasets are not challenging enough. For example, two matching claims ``London is the capital city of the UK'' and ``UK's capital is London'' can be easily identified as a paraphrase of each other. However, matching claims are not necessarily paraphrases of each other, e.g. ``The president of the USA is Joe Biden'' and ``The president of the United States is Barack Obama''. The former is correct at the time of writing, whereas the latter is not correct, while they are matching claims because they can both be fact-checked with the same piece of evidence.

As an avenue for future work, to enhance model explainability, chain-of-thought methods can be implemented to let a model decompose the input and better understand nested sub-claims. Combinations of prompts could be also further examined.

\section*{Limitations}

The results depend on the choice of prompt templates and their formulations on surface level. Use of hand-crafted templates depends on the researcher's intuition, leading to limitations in the experimentation and automating the prompt generation that is left for future work. Likewise, choice of few-shot train examples is also important. Alternative prompt-learning and -engineering methods should be implemented, automatically searching for best prompts and templates. Different LLMs require different prompt styles/selection of examples/prompt formatting. More fine-grained techniques, depending on a particular LLM, could improve their results and help understand why different prompt templates work better for different models. Fine-tuning approaches to LLMs, with more training data, could also reduce in the future the impact of the train examples selection. More detailed analysis of model error types is also the focus of further research. Using ``ensemble'' instructions, as a method, should be further investigated. On short texts, models in their explanations demonstrate that they mostly consider both system and user instruction combined. 

Reproducibility of classification results, with the same LLM parameters, remains a challenging issue. But, for example, for five runs of GPT-3.5 and Gemini with PD-6 template, the results differed only slightly: mean F1 scores 84.7\% and 90.5\%, with standard error 0.25 and 0.10, respectively. The borderline examples are ambiguous, despite the gold label about their match. 

Our experiments were conducted  with English data only, which could be further extended to other languages using multilingual LLMs. CM is a new task, and only limited datasets are available. We used all two existing datasets, as a basis for our research. Dataset from~\citep{nakov-clef-2022} contains English and Arabic parts. Dataset from~\citep{kazemi2022} includes English, Hindi, Spanish, and Portuguese. We provide the first study on our CM approach and use only English parts, for prompt engineering and evaluation purposes, as we do not have native speakers in the aforementioned languages. There is indeed need to create more datasets, and we are planning to create more CM datasets, including the multilingual ones.

% Bibliography entries for the entire Anthology, followed by custom entries
%\bibliography{anthology,custom}
% Custom bibliography entries only
\bibliography{custom}

\appendix

\section{Appendix}
\label{sec:appendix}

Table \ref{tab:claimmatch} shows 5 matching and 5 non-matching claim pairs from ClaimMatch, whereas Table \ref{tab:trainex} shows 10 randomly selected claim pairs used for few-shot training.

\begin{table*}[htb]
    \centering
    \small
    \begin{tabular}{p{0.43\linewidth} | p{0.43\linewidth} | p{0.1\linewidth}} 
     \hline
     \textbf{Claim 1} & \textbf{Claim 2} & Class label \\    
     \hline
     1. Here is a homeless, elderly couple from California sleeping on concrete with no water from a toilet or even a toilet at all. If only they were illegals, the Democrats might actually care about their well-being — derek schwartz (derek\_mafs) July 13, 2019 & Is This an ‘Elderly Homeless Couple’ in California? Viral social media posts claimed Democratic lawmakers might care about the couple's well-being if they were "illegals.". A photograph depicts an elderly homeless couple sleeping on the concrete in California. & Positive \\
     \cdashline{1-3}
     2. netflix what is this madness?? the literal source of my happiness, a.k.a. Friends, is leaving?? heartbroken — Noelle Michaud (noellemichaud14) December 2, 2018 & Did Netflix Announce They Will Be Removing ‘Friends’ from Their Streaming Catalog? The online streaming service caused consternation in December 2018 when they briefly signalled the hit 90's sitcom would no longer be available after 1 January 2019. Netflix announced that 'Friends' would no longer be available for customers to stream after the end of 2018. & Positive \\
     \cdashline{1-3}
     3. The Sheriff in Union County, Arkansas is putting Nike t-shirts on people they arrest and making them wear them during mugshots. Source says it is to mock Nike and Colin Kaepernick. Disgusting. — Shaun King (shaunking) October 11, 2018 & Did a Sheriff’s Dept. Dress Arrestees in Nike Shirts for Their Mugshots? Was the use of Nike shirts for mugshots a subtle jab at the company's political choices, or a practical use of available material? The Union County Sheriff's Department in Arkansas dressed numerous arrestees in Nike shirts for their mugshots as a jab at the company. & Positive \\
     \cdashline{1-3}
     4. Should we even be surprised anymore? Peter Strzok’s sister-in-law works with Christine Blasey Ford’s brother. (Corrected version with Jill Strzok) — PaleoHorse (PaleoHorse) September 26, 2018 & Did Christine Blasey Ford’s Brother Work with a Close Relative of Peter Strzok’s? Conspiracy theorists claim they've found 'three degrees of separation' between fired FBI agent Peter Strzok and Brett Kavanaugh's sexual assault accuser Christine Blasey Ford. Christine Blasey Ford's brother, Tom Blasey, worked in the same company as a close relative (wife, sister, or sister-in-law) of former FBI agent Peter Strzok. & Positive \\
     \cdashline{1-3}
     5. Shocking video from 2017 shows NancyPelosi describing the Democrat attack plan called the “Wrap-up Smear” (the technique used against Justice Brett Kavanaugh ) movingUSforward — THE SCOOP (TheScoop\_US) October 8, 2018 & Did Nancy Pelosi Admit Democrats Use a Tactic Called the ‘Wrap-Up Smear?’. Unreliable sources claimed Pelosi admitted on video that Democrats use such a tactic, but in reality she ascribed it to Republicans. In a C-SPAN video, U.S. House Minority Leader Nancy Pelosi revealed that Democrats use a political smear tactic she called the "wrap-up smear." & Positive \\
     \cdashline{1-3}
     6. Polls are starting to look really bad for Obama. Looks like he’ll have to start a war or major conflict to win. Don’t put it past him! — Donald J. Trump (realDonaldTrump) October 17, 2012 & Is Amazon Bankrupting the United States Postal Service? President Donald Trump has claimed the giant online retailing company is jeopardizing USPS's financial health. Amazon's use of the USPS for package delivery services has resulted in a severe financial loss to the Postal Service. & Negative \\
     \cdashline{1-3}
     7. Suspended by facebook. For blowing the whistle. On something they have known privately for 2 years. — Christopher Wylie (chrisinsilico) March 18, 2018 & Did George Soros Pay ‘March for Our Lives’ Protesters \$300 Each? As hundreds of thousands of students and supporters took to the streets to protest gun violence, conspiracy theorists blamed a familiar bogeyman. Organizers working for billionaire George Soros ran Craigslist ads offering \$300 each to individuals participating in the 24 March 2018 "March for Our Lives" protests. & Negative \\
     \cdashline{1-3}
     8. Earlier, Planned Parenthood suggested ‘we need a Disney princess who’s had an abortion’ and then they deleted it. So, in case you missed it: — Chet Cannon (Chet\_Cannon) March 27, 2018 & Did an Instagram Post by Rihanna Cause Snapchat’s Stock Market Value to Fall? The megastar's exhortation to her 61 million followers to "throw away" the app likely did make a significant contribution to a \$750 million one-day loss. Rihanna's Instagram message to followers to throw away the Snapchat app caused the company's share value to fall by hundreds of millions of dollars in one day. & Negative \\
     \cdashline{1-3}
     9. Yeti cuts ties with the NRA Foundation — Mark R. Levin (marklevinshow) April 23, 2018 & Did Donald Trump Tweet Warnings About Obama Ordering Syrian Airstrikes? In 2013, Donald Trump issued multiple warnings to his predecessor about attacking Syria without congressional approval. Donald Trump leveled the same criticisms against President Barack Obama over Syrian strikes that were later used against him. & Negative \\
     \cdashline{1-3}
     10. Today we are flipping our iconic McDonalds logo in honour of women everywhere. IWD2018 — Steve Easterbrook (SteveEasterbrk) March 8, 2018 & Does Delta Airlines Give Planned Parenthood Members Discounted Rates? A Georgia senator has claimed — without any supporting evidence — that Delta offers discounts on air travel to Planned Parenthood members. Delta Air Lines gives members of Planned Parenthood discounted rates on air travel. & Negative \\
    \hline
    \end{tabular}
    \caption{Positive and negative class examples from the ClaimMatch dataset.}
    \label{tab:claimmatch}
\end{table*}

\begin{table*}[htb]
    \centering
    \small
    \begin{tabular}{p{0.43\linewidth} | p{0.43\linewidth} | p{0.1\linewidth}} 
     \hline
     \textbf{Claim 1} & \textbf{Claim 2} & Class label \\    
     \hline
     1. Statement 1: How are butterflies surviving the AustralianFires? Julie Favell was putting out water for wildlife that survived the fires when she witnessed common brown butterflies (Heteronympha merope) fluttering in a moist wombat hole. Footage by Julie Favell — Center for Bio Div (CenterForBioDiv) January 14, 2020 & Statement 2: Are Wombats Inviting Animals Into Their Burrows to Escape Australia Fires? Gather 'round to hear the tale of the wombat hero ... or at least the wombat's big burrow. Wombats are herding animals and inviting them into their burrows in order to escape the wildfires in Australia. & Positive \\
     \cdashline{1-3}
     2. Statement 1: Trump needs to immediately divest from his businesses and comply with the emoluments clause. Iran could threaten Trump hotels *worldwide* and he could provoke war over the loss of revenue from skittish guests. His business interests should not be driving military decisions. — Ilhan Omar (IlhanMN) January 6, 2020 & Statement 2: No, U.S. Rep. Ilhan Omar Didn’t Give ‘Treasonous’ Military Advice to Iran. An article by the anti-Muslim activist Robert Spencer prompted threats and incitements of violence and murder against the Minnesota congresswoman in January 2020. In January 2020, U.S. Rep. Ilhan Omar advised Iran to attack Trump-branded hotels in the world, thus committing treason. & Positive \\ 
     \cdashline{1-3}
     3. Statement 1: Once more for the folks who missed the tweet the first time I posted it. Minecraft is NOT stopping, Mojang is NOT closing. Minecraft — Helen Z PAXSouth (HelenAngel) January 2, 2020 & Statement 2: Did Trump Ask Advisers About ‘Nuking’ Hurricanes? A report from Axios suggested the question was raised on more than one occasion. U.S. President Donald Trump has asked his advisers about the feasibility of stopping hurricanes with nuclear bombs. & Negative \\ 
     \cdashline{1-3}
     4. Statement 1: Earlier today, a news outlet accurately reported that a subcontractor for one of our vendors was using prison workers to make phone calls on behalf of my campaign. After learning this, we immediately ended our relationship with that company. Full statement below: — Mike Bloomberg (MikeBloomberg) December 24, 2019 & Statement 2: Is Minecraft Shutting Down? Internet pranks are not confined to April Fools' Day. The popular video game Minecraft is shutting down in 2020. & Negative \\
     \cdashline{1-3}
     5. Statement 1: A number of fraudulent text messages informing individuals they have been selected for a military draft have circulated throughout the country this week. & Statement 2: Is US Army Sending Texts About a Military Draft? The military has been an all-volunteer force since 1973. The U.S. Army is sending text messages informing people they've been selected for the military draft. & Positive \\
     \cdashline{1-3}
     6. Statement 1: This is the third spill along the Keystone pipeline’s route in less than three years. Pipeline projects like this will continue to spill and pollute the earth until we shut them down for good and replace with 100\% clean energy. Zeroto100 — Earthjustice (Earthjustice) October 31, 2019 & Statement 2: Did Michael Bloomberg’s Presidential Campaign Use Prison Labor? The former New York mayor landed in hot water over a vendor's reported use of prison workers to place calls to would-be voters. In 2019, the presidential campaign of Michael Bloomberg worked with a vendor that used prison labor on the campaign's behalf. & Negative \\
     \cdashline{1-3}
     7. Statement 1: America has more governors who’ve worn blackface than black governors. — Samuel Sinyangwe (samswey) August 30, 2019 & Statement 2: Was the Keystone Pipeline Shut Down After Leaking Oil? The pipeline did leak in North Dakota, but claims by alternative media that "no one" was talking about it were demonstrably false. The Keystone Pipeline in North Dakota was shut down after it leaked oil. & Negative \\
     \cdashline{1-3}
     8. Statement 1: Hurricane Dorian washed up bricks of cocaine on Florida’s coast — NowThis (nowthisnews) September 10, 2019 & Statement 2: Does America Have More Governors Who Have Worn Blackface Than Black Governors? The question gets at the heart of whether an African American currently leads the executive branch of any state in the United States. The U.S. has more governors who have worn blackface than actual black governors. & Negative \\
     \cdashline{1-3}
     9. Statement 1: The US drone attack on Soleimani caught on camera. IranUsa — Olaudah Equiano (RealOlaudah) January 6, 2020 & Statement 2: Does This Video Show the Drone Strike That Killed Soleimani? Iran Gen. Qassem Soleimani was killed by a U.S. drone strike in Baghdad on Jan. 3, 2020. A video shows the U.S.-ordered drone strike that killed Iran Gen. Qassem Soleimani. & Positive \\
     \cdashline{1-3}
     10. Statement 1: Picture of engagement rings removed from the fingers of World War- II victims. Imagine how many love stories were buried!Wars are never easy!Nd we r making fun of it.. shameful act.. WWIIl WW3 — ChaudhRy Saab (tera\_Lover\_) January 5, 2020 & Statement 2: Does This Photo Show Wedding Rings Taken from Holocaust Victims? “Every wedding ring here represents a home broken and a human murdered by the Germans.". A photograph documents a cache of wedding rings removed from Holocaust victims. & Positive \\
    \hline
    \end{tabular}
    \caption{Few-shot train examples.}
    \label{tab:trainex}
\end{table*} 

\end{document}